\documentclass{article}
\usepackage{spconf,amsmath,graphicx}
\usepackage{subfigure}

\usepackage{algorithm}
\usepackage{algorithmic}
\usepackage[mathscr]{eucal}
\usepackage{amsmath}
\usepackage{amsfonts}
\usepackage{multirow}
\usepackage{xcolor}
\usepackage{newfloat}
\usepackage{listings}
\usepackage{booktabs}
\usepackage{url}
\usepackage{enumitem}

\usepackage{hyperref}
\usepackage{url}

\hypersetup{
    colorlinks=true,
    linkcolor=blue,
    citecolor=blue,
    urlcolor=blue
}

\title{Self-Adaptive Scale Handling for Forecasting Time Series with Scale Heterogeneity}
\name{Xu Zhang\textsuperscript{1}\thanks{This work was conducted when Xu Zhang and Yunzhi Wu was interning at Ant Group. This work was supported by Ant Group Research Fund. *Corresponding author.
}, Zhengang Huang\textsuperscript{2},Yunzhi Wu\textsuperscript{1}, Xun Lu\textsuperscript{2}, Erpeng Qi\textsuperscript{2}, Yunkai Chen\textsuperscript{2}, Zhongya Xue\textsuperscript{2}, \\
Peng Wang\textsuperscript{1*}, Wei Wang\textsuperscript{1}}
\address{\textsuperscript{1}College of Computer Science and Artificial Intelligence, Fudan University, Shanghai, China\\
\textsuperscript{2}Ant Group, Shanghai, China}

\begin{document}
%\ninept
%
\maketitle
\begin{abstract}
Current time series forecasting (TSF) research predominantly focuses on scale-homogeneous data, where different time series share similar numerical magnitude ranges. However, in real-world industrial scenarios such as financial product sales, different time series often differ by orders of magnitude (scale heterogeneity). Since these series share similar temporal patterns, joint modeling is desirable for better data utilization, yet existing scaling methods either compress low-scale signals (global normalization) or destroy semantic discriminability and amplify inverse-scaling errors (window-based scaling). This paper proposes a self-Adaptive Scale-handling (AS) module that learns adaptive scale factors tailored to each input, preserving semantic discriminability while reducing inverse-scaling errors. AS consists of Scale Calibrating (SC), which calibrates prior mean scaling factors through neural networks, and Scaling Selection (SS), which decides whether to apply calibration or retain the original factor, avoiding over-calibration. Experiments on real-world fund sales datasets from Ant Fortune and Alipay show that AS seamlessly integrates into popular TSF models and consistently improves their performance. The code and dataset are available at \url{https://github.com/Meteor-Stars/ASTSF}.
\end{abstract}

\begin{figure*}[h!]
% %\vspace{-0.25cm}
\centerline{\includegraphics[width=\linewidth]{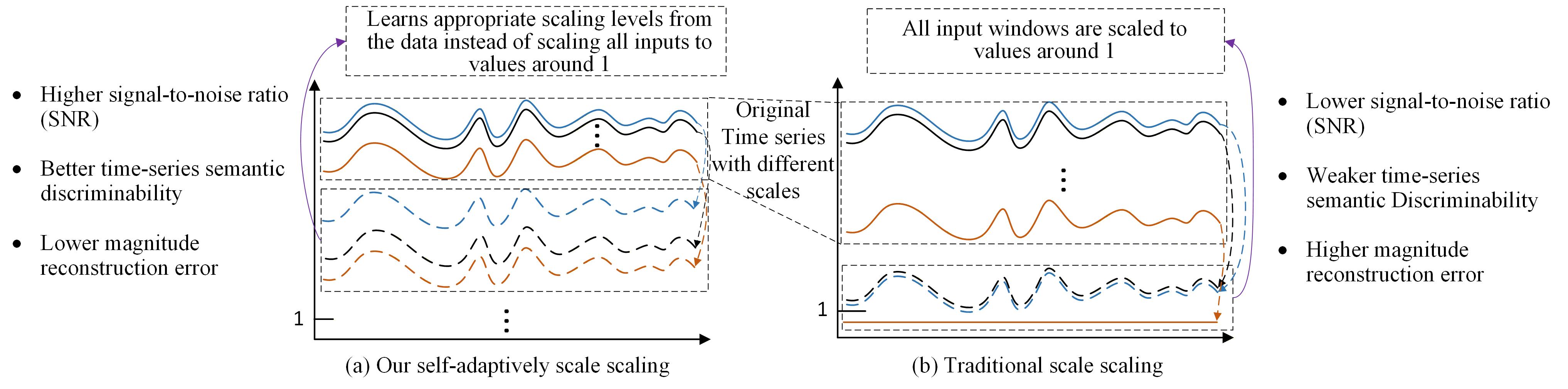} }

\caption{Comparison between our adaptive scale scaling (left) and traditional scale scaling (right). Our method learns appropriate scaling levels from the data instead of uniformly scaling all inputs to around 1, resulting in higher signal-to-noise ratio, better time-series semantic discriminability, and lower magnitude reconstruction error.}
\label{fig:scale_scaling_moti}

\end{figure*}

\section{Introduction}
Time series forecasting (TSF) is essential in many real-world applications, including weather prediction~\cite{angryk2020multivariate_intr_weather}, traffic flow estimation~\cite{chen2001freeway_traffic} and  financial inventory planning~\cite{zhang2025multi}. Based on the \textbf{scale} (i.e., \textbf{numerical magnitude level}) distribution properties of the data, TSF tasks can be categorized into \textit{scale-homogeneous} and \textit{scale-heterogeneous} settings. Scale-homogeneous data exhibits relatively small variations in numerical scales, such as temperature forecasting. Current TS research predominantly focuses on scale-homogeneous settings~\cite{li2019enhancing,zhang2026lost,zhou2021informer,zhou2022fedformer,zhang2026amortized,zeng2023transformers_effect,Zhang2025_GFEF,zhou2022film,ts_mixer_1,ts_mix_2,zhang2026diff,wu2021autoformer}, while scale-heterogeneous scenarios remain underexplored. In contrast, scale-heterogeneous time series are prevalent in industrial scenarios. For instance, sales volumes of different fund products on financial platforms or different items on e-commerce platforms can differ by up to three orders of magnitude.

In practice, scale-heterogeneous time series may share similar temporal patterns (e.g., seasonal trends across different fund products), which motivates jointly constructing samples across variables to improve data utilization and forecasting accuracy, especially for variables with limited data. However, jointly training across different scales introduces the heterogeneity challenge. On one hand, models tend to overfit to a specific scale range and fail to generalize to data at other scales. On the other hand, when using global standardization or normalization~\cite{zhou2021informer,wu2021autoformer}, the large scale gap between high- and low-scale samples causes low-scale values to be excessively compressed after transformation, reducing their signal-to-noise ratio and degrading gradient updates during training. An intuitive alternative is variable-specific modeling, which assigns independent parameters per series, but ignores shared cross-variable patterns, cannot address intra-variable scale shifts over time, and scales poorly with the number of variables. A common alternative is window-based scaling~\cite{salinas2020deepar}, which divides each sliding window by its mean to normalize values to around 1. As illustrated in Figure~\ref{fig:scale_scaling_moti} (right), this uniform scaling has two critical drawbacks: (1) it destroys \textit{semantic discriminability}, meaning the model can no longer perceive that one series is inherently orders of magnitude larger than another; (2) during inverse scaling, high-scale series with large scale factors amplify small normalized-space errors into substantial absolute errors.

To address these limitations, we propose a self-\textbf{A}daptive \textbf{S}cale-handling (AS) module. As shown in Figure~\ref{fig:scale_scaling_moti} (left), instead of uniformly mapping all windows to around 1, our AS module learns adaptive scale factors tailored to each input, preserving semantic discriminability while reducing inverse-scaling errors. The module consists of two sub-modules: (1) a \textbf{Scale Calibrating (SC)} sub-module that calibrates priori scale factors through neural networks, and (2) a \textbf{Scaling Selection (SS)} sub-module that decides, for each window, whether to use the calibrated scale factor or retain the original priori factor, avoiding harmful over-calibration. The AS module can be seamlessly integrated into any TSF model for end-to-end training. Our contributions are summarized as follows:

\begin{itemize}
    \item We propose the Adaptive Scale-handling (AS) module for scale-heterogeneous TSF. The Scale Calibrating sub-module learns to calibrate prior mean scaling factors, reducing inverse-scaling restoration errors while preserving scale semantics.
    \item Through parameterizing a Bernoulli distribution via Gumbel-Softmax, we propose the Scaling Selection sub-module, enabling the model to autonomously decide whether to calibrate or retain the original prior mean scaling factor for each input, preventing performance degradation from unnecessary calibration.
    \item We conduct comprehensive experiments on collected real-world industrial datasets, combining AS with popular TSF models. Results demonstrate consistent improvements, and we provide a systematic study of training strategies, loss functions, and data processing methods for scale-heterogeneous TSF.
\end{itemize}

\section{Related Work}

\textbf{Time series normalization and scaling.}
Most TSF methods employ global standardization or normalization as preprocessing~\cite{zhou2021informer,wu2021autoformer}, assuming data follows a roughly homogeneous distribution. For scale-heterogeneous data, window-based scaling~\cite{salinas2020deepar} offers a practical alternative by dividing each window by its local mean. While this partially alleviates cross-series scale differences, it forces all windows to a uniform range regardless of their original scale, weakening inter-series discriminability and introducing large reconstruction errors during inverse scaling, which motivates our adaptive approach.

\textbf{Time series forecasting.}
Deep learning-based approaches have become dominant in time series forecasting and can be broadly categorized into Transformer-based models and lightweight linear architectures. Transformer-based methods leverage strong sequence modeling capabilities: Informer~\cite{zhou2021informer} introduces ProbSparse attention to reduce complexity for long sequences, Autoformer~\cite{wu2021autoformer} incorporates series decomposition with auto-correlation mechanisms, Performer~\cite{performer} approximates full attention via random feature maps for linear-time computation, and MLF~\cite{zhang2025multi} introduces multi-period modeling with adaptive patching and weighted fusion to capture multiscale temporal dynamics. Meanwhile, lightweight linear architectures demonstrate that simple designs can achieve competitive performance with high efficiency. LSINet~\cite{zhang2025lightweight} models temporal dynamics through a multi-head sparse interaction mechanism, while SEMixer~\cite{zhang2026semixer} captures temporal dependencies via random attention and progressive multi-scale mixing strategies.

\section{Method}
%\vspace{-0.1cm}
\subsection{Problem Formulation}
\begin{figure*}[h]
% %\vspace{-0.25cm}
\centerline{\includegraphics[width=0.85\linewidth]{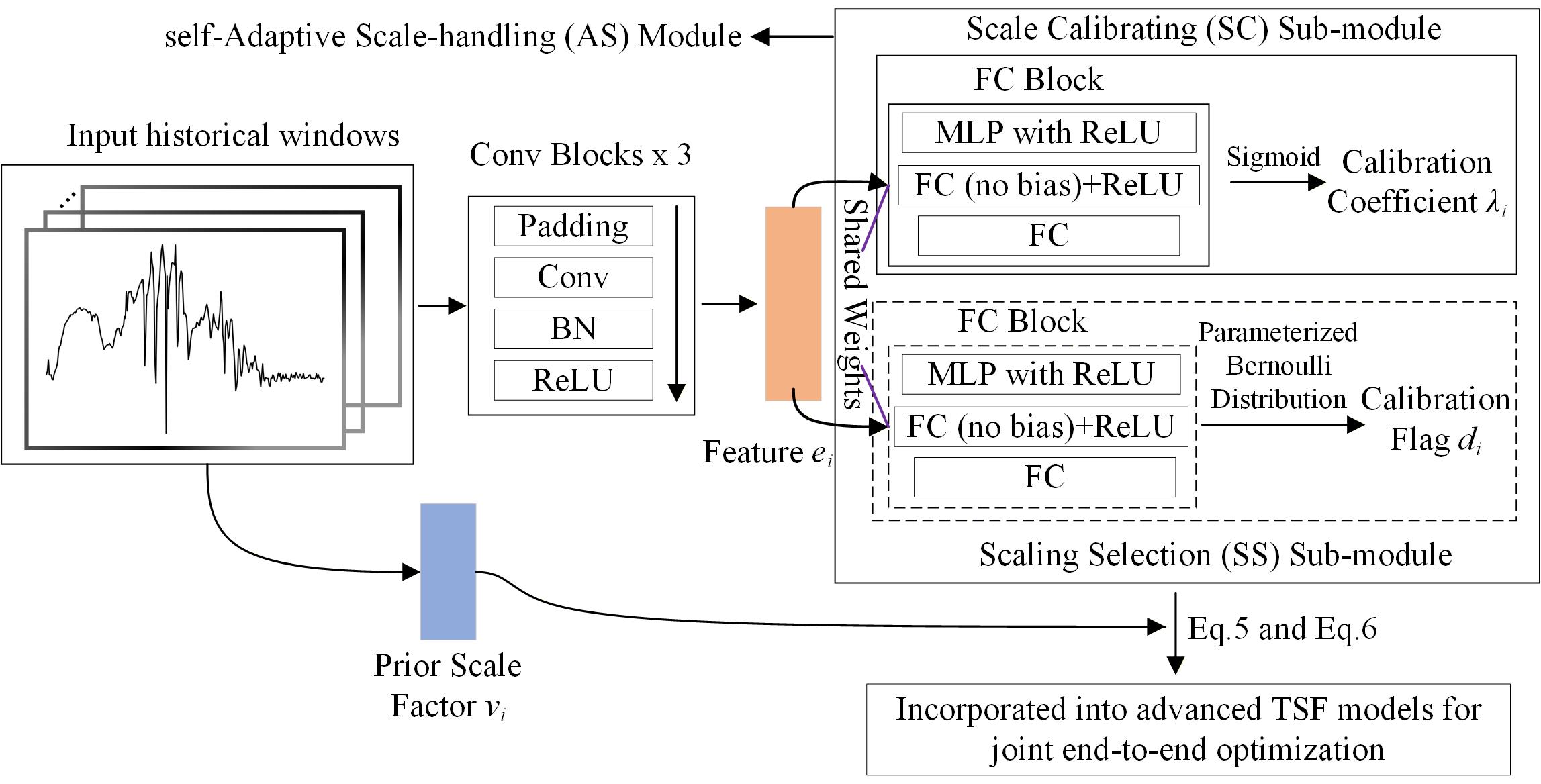} }
%\vspace{-0.3cm}
\caption{Framework for forecasting TS with scale heterogeneity based on the proposed self-Adaptive Scale-handling module. AS consists of three steps:
(1) extracting temporal features from input time series windows;
(2) computing the Calibration Coefficient $\hat{\lambda}$ and a binary Calibration Flag $d_i$;
(3) applying scaling to the prior mean scaling factor $v_i$. Specifically, if the Calibration Flag is 1, the prior mean scaling factor is further calibrated; otherwise, the original prior mean scaling factor is retained without modification.}
\label{fig:Framework}
%\vspace{-0.6cm}
\end{figure*}

Given a historical input window (multivariate time series instance) $\mathcal{X}^h=[x_1,x_2,...,x_n] \in \mathbb {R}^{n \times c}$ with the length of $n$, time series forecasting (TSF) tasks aim to forecast the future $m$ steps $\mathcal{X}^f=[x_{n+1},x_{n+2},...,x_{n+m}] \in \mathbb {R}^{m \times c}$ for all $c$ variables. In scale-heterogeneous scenarios, multiple variables exhibiting similar temporal patterns but vastly different numerical scales are jointly modeled by a shared forecasting backbone. We propose the AS module to adaptively handle scale diversity, enabling the model to effectively process time series data at various scales. The overall framework is shown in Figure~\ref{fig:Framework}.

\subsection{Scale Calibrating (SC Sub-module)}
\textbf{Prior Mean Scaling Factors.} Through sliding windows, we can obtain pairs of multiple historical input windows (single is $\mathcal{X}^h$) and future to-be-forecasted windows (single is $\mathcal{X}^f$). We take a single one to describe our method for simplicity. Similar to DeepAR~\cite{salinas2020deepar}, we can obtain the scale factor $v_i$ by:
%\vspace{-0.2cm}
 \begin{equation}
\label{equ:factor_v}
v_i=1+\frac{1}{\hat{t}} \sum_{t=1}^{n} \mathcal{X}^h_{i,t}
%\vspace{-0.2cm}
\end{equation}
where $\hat{t}$ is the number of non-zero time steps. $i$ is the sample index. $1$ is to ensure numerical stability.

\textbf{Learning Calibrated Scale Factors.} $v_i$ can scale all input windows to around 1, helping the model deal with TS data at various scales. However, this may lead to the model's inability to distinguish differences within the various scales and larger scale restoration errors when forecasting TS data with larger $v_i$. Hence, we can enhance the forecasting performance by learning an adaptive scale factor that not only can help the model relieve scale heterogeneity but also increase TS distinguishability and reduce scale restoration errors. We achieve this through deep learning and data-driven approaches. Specifically, we first use 3 convolution blocks to obtain the feature $e_i$ from the input window. $\mathcal{X}^h_i$ after using the padding function ($Pad$) is sent to a convolutional neural network ($Conv$) and then followed by a batch normalization layer ($BN$) and $ReLU$ function. The formulation of a single block is:
%\vspace{-0.15cm}
 \begin{equation}
\label{equ:features_e}
e_i=ReLU(BN(Conv(Pad(\mathcal{X}^h_i))))
%\vspace{-0.15cm}
\end{equation}
It is difficult to force a model to directly learn the accurate scale factors due to the significant scale differences. Hence, we use a fully connected (FC) block followed by a $Sigmoid$ function to learn the calibrated coefficients $\lambda$ for the scale factors instead of directly learning them. The FC block consists of 5 FC layers. The first three FC layers form an $MLP$ network, the fourth discards the learnable bias parameters, and the fifth linear layer is equipped with a sigmoid function and obtains the learned calibrated coefficients. Each FC layer in the $MLP$ is followed by a ReLU function. Formulation is:
%\vspace{-0.15cm}
 \begin{equation}
\label{equ:coefficients}
\lambda_i=\sigma(\Gamma(e_i)), \Gamma=FC_5(ReLU(FC_4(MLP(\cdot))))
%\vspace{-0.15cm}
\end{equation}
Where $\sigma$ denotes $Sigmoid$ function. In practice, we average $\lambda_i$ over the batch to obtain $\hat{\lambda}$ for stable training and testing. The calibrated scale factors are then $\overline{v}_i=\hat{\lambda} \times v_i$.
%\vspace{-0.2cm}
\subsection{Scaling Selection (SS Sub-module)}
Eq.~\ref{equ:coefficients} forces the model to learn calibrated coefficients for all variables in $\mathcal{X}^h_i$. This can lead to some initially well-performing factors being over-calibrated, resulting in a decrease in performance. Hence, we further propose a scaling selection (SS) module to keep the well-performing factors and calibrate the others, avoiding unnecessary performance degradation. 
The SS module depends on the sampling results of the binary Bernoulli distribution, i.e., whether to use the calibrated or original prior mean scaling factor. However, unlike classification tasks where discrete labels do not participate in gradient computation, the discrete sampling variable here directly affects the forward pass and must be differentiable. Hence, we use Gumbel-Softmax distribution and Gumbel reparameterization trick~\cite{Categorical_gumbel,Concerate} to make the sampling process differentiable. Specifically, a separate FC block $\Phi$ (with the same architecture as $\Gamma$ but independent parameters) maps the feature $e_i$ to Bernoulli probabilities.
We can obtain the sampling result $d_i$:
%\vspace{-0.1cm}
 \begin{equation}
 \label{equ:decision_formula}
  d_i = \frac{exp((log \varphi^c_i+g^c_i)/\tau)}{\sum exp((log \varphi_i+g_i)/\tau) },\varphi^c_i=\Phi(e_i)
%\vspace{-0.1cm}
\end{equation}
where $g^c_i=-log(-log(u))$ is sampled from the Gumbel Softmax distribution through inverse transform sampling of $u\sim$ Uniform(0,1), $c \in \{0,1\}$ denotes whether to conduct scaling on the current sample. $\tau$ is the temperature that controls the smoothness of the Gumbel-Softmax distribution. 

Now, we can optimize the model parameters through gradient descent to obtain the good $d_i$. It contains the probability of whether to conduct scale calibrating operation on the current input window$\mathcal{X}^h_i$. Hence, we can selectively perform scale calibrating on scale factors $v_i$ to obtain better model performance in a data-driven manner based on $d_i$:
\begin{equation}
%\vspace{-0.2cm}
\label{equ:final_scale_factors}
    \hat{v}_{i}=
    \begin{cases}
    v_i \times \hat{\lambda} &  \text{if} \ \mathop{\arg\max}(d_i)=1 \\
    v_i &  \text{otherwise} \\
    \end{cases}
%\vspace{-0.1cm}
\end{equation}
During the forecasting process, the input windows $X^h$ are first scaled by $\hat{v}_{i}$. Then, the forecasting results $\hat{X}^f$ based on TSF models ($f(\cdot)$) are inversely scaled:
%\vspace{-0.15cm}
 \begin{equation}
 \label{equ:prediction}
   \hat{\mathcal{X}}^f=f(\mathcal{X}^h/\hat{v}_i) \times \hat{v}_i
%\vspace{-0.15cm}
\end{equation}
Note that $\hat{v}_i$ only works on the instance level, denoting that all time steps in $\mathcal{X}^h_i$ will be scaled by the same $\hat{v}_i$.

\section{EXPERIMENTS}
\subsection{Experimental Settings}
\textbf{Dataset.} We collect fund sales datasets from Ant Fortune, which is an online wealth management platform on the Alipay APP. They are divided into two groups based on the holding period for comprehensive experiment evaluation, called Fund$_1$ (66 fund sales datasets) and Fund$_2$ (106 fund sales datasets). The sales of different fund products exhibit scale heterogeneity and are suitable to validate our method. The fund dataset has applying and redeeming amounts to be forecasted. Different products are used to jointly construct samples based on sliding windows. All datasets are divided into training, validation, and testing set as the ratio of 7:1:2.

\textbf{Baselines and Implementation Details.} We combine our method with popular TSF models Informer~\cite{zhou2021informer}, vanilla Transformer~\cite{att_need}, Performer~\cite{performer} and Autoformer~\cite{wu2021autoformer}. All models have two variants: model only equipped with SC module or both equipped with SC and SS modules.  We report the better result of them. For transformer baselines, the number of layers in the encoder is fixed at 4, the number of heads and $d_{model}$ in the self-attention mechanism is fixed at 8 and 512, and the hidden size of the feed forward layer is fixed at 2048. We use TS with sequence length 30 to forecast future 5 and 10 time steps.  All experiments in this study are conducted on NVIDIA GeForce RTX 3090 GPU on PyTorch.

\textbf{Evaluation Metrics.} To avoid the influence of scale heterogeneity, we use Weighted Mean Absolute Percentage Error (WMAPE, WMA. for short), Root Squared Error (RSE), and Percentage Mean Absolute Error (PMAE) as metrics, and Lower is better. WMA. contains the sum errors of both applying and redeeming amounts. Unless otherwise specified, all models are trained on WMAPE loss for better performance.

 \begin{table}[h!]
    \setlength{\tabcolsep}{1.3pt}
    % {|>{\setlength{\tabcolsep}{3pt}}c|c|c|}
    \centering
    \caption{Comparison of TSF models under different scale-handling strategies: our Adaptive Scale-handling (``AS''), vanilla scaling (``VS'', using prior mean scaling factor $v_i$), and no preprocessing (``nop''). The last two rows additionally compare against global standardization (``std'') and normalization (``nor''). Lower is better.  }
    %\vspace{0.2cm}
    \label{tab:main_res}
    % \begin{tabular}{c|c|p{20pt}p{20pt}|cc|cc|cc|cc|cc}
    \begin{tabular}{c|c|cc|cc|cc}
        \hline
        \multirow{2}{*}{\shortstack{Methods/\\Datasets}} & &  \multicolumn{2}{c|}{Autoformer$_{AS}$} &  \multicolumn{2}{c|}{Autoformer$_{VS}$}&  \multicolumn{2}{c}{Autoformer$_{nop}$} \\
         & & WMA. & RSE & WMA. & RSE & WMA. & RSE    \\ 
        \midrule[1pt]
        \multirow{2}{*}{Fund$_1$} &5 &\textbf{93.27} &\textbf{60.38}&97.02 &60.78&106.01 &66.77 \\
         & 10 &\textbf{93.78} &\textbf{61.32}&96.12 &61.81&106.06 &67.92 \\
         \midrule[1pt]
         \multirow{2}{*}{Fund$_2$} &5 &\textbf{86.33} &\textbf{59.93}&89.64 &60.84&98.49 &67.00 \\
         & 10 &\textbf{88.96} &\textbf{61.46}&92.07 &62.57&98.28 &68.55 \\

        \hline
        \multirow{2}{*}{\shortstack{Methods/\\Datasets}} & &  \multicolumn{2}{c|}{Transformer$_{AS}$} &  \multicolumn{2}{c|}{Transformer$_{VS}$}&  \multicolumn{2}{c}{Transformer$_{nop}$} \\
        & & WMA. & RSE & WMA. & RSE & WMA. & RSE    \\ 
        \midrule[1pt]
        \multirow{2}{*}{Fund$_1$} &5 &\textbf{91.75} &\textbf{62.27}&93.62. &64.30&102.08 &74.61 \\
         & 10 &\textbf{94.64} &\textbf{62.42}&96.34 &64.24&105.30 &75.93  \\
         \midrule[1pt]
        \multirow{2}{*}{Fund$_2$} &5 &\textbf{80.12} &\textbf{59.81}&80.94 &61.19&86.89 &70.97 \\
         & 10 &\textbf{85.50} &\textbf{62.05}&86.16 &62.37&92.57 &73.38 \\
        \hline
        \multirow{2}{*}{\shortstack{Methods/\\Datasets}} & &  \multicolumn{2}{c|}{Performer$_{AS}$} &  \multicolumn{2}{c|}{Performer$_{VS}$}&  \multicolumn{2}{c}{Performer$_{nop}$} \\
        & & WMA. & RSE & WMA. & RSE & WMA. & RSE    \\ 
        \midrule[1pt]
        \multirow{2}{*}{Fund$_1$} &5 &\textbf{89.12} &\textbf{59.55}&90.72 &60.84 &105.13 &76.19 \\
         & 10 &\textbf{93.15} &\textbf{60.82}&94.33 &61.63&111.04 &78.19  \\
         \midrule[1pt]
        \multirow{2}{*}{Fund$_2$} &5 &\textbf{79.18} &\textbf{58.23}&81.22 &59.86&88.84 &71.66 \\
         & 10 &\textbf{86.13} &\textbf{61.36}&86.64 &61.47&95.62 &74.66 \\
        % \midrule[1pt]
        
        \hline
        \multirow{2}{*}{\shortstack{Methods/\\Datasets}} & &  \multicolumn{2}{c|}{Informer$_{AS}$} &  \multicolumn{2}{c|}{Informer$_{VS}$}&  \multicolumn{2}{c}{Informer$_{nop}$} \\
        & & WMA. & RSE & WMA. & RSE & WMA. & RSE    \\ 
        \midrule[1pt]
        \multirow{2}{*}{Fund$_1$} &5 &\textbf{83.07} &\textbf{59.43}&83.53 &60.55&97.21 &74.18 \\
         & 10 &\textbf{94.01} &\textbf{64.08}&94.86 &65.35&108.85 &78.63  \\
         \midrule[1pt]
        \multirow{2}{*}{Fund$_2$} &5 &\textbf{75.56} &57.19&75.95 &\textbf{57.83}&85.74 &70.59 \\
         & 10 &\textbf{83.73} &\textbf{61.26}&84.69 &62.53&94.07 &74.13 \\

        \hline
        \multirow{2}{*}{\shortstack{Methods/\\Datasets}} & &  \multicolumn{2}{c|}{Performer$_{nop}$} &  \multicolumn{2}{c}{Performer$_{std}$}&  \multicolumn{2}{c}{Performer$_{nor}$} \\
         & & WMA. & RSE & WMA. & RSE & WMA. & RSE    \\ 
        \midrule[1pt]
        \multirow{1}{*}{Fund$_1$} 
         & 10 &111.04 &78.19&114.76 &88.62&114.48 &88.62  \\
        \midrule[1pt]
        \multirow{1}{*}{Fund$_2$} 
         & 10 &95.62 &74.66&119.51 &88.57&117.77 &88.57  \\
        \hline
    \end{tabular}
%\vspace{-0.4cm}
\end{table}

 \begin{table}[h!]
    \setlength{\tabcolsep}{3pt}
    % {|>{\setlength{\tabcolsep}{3pt}}c|c|c|}
    \centering
    \caption{Ablation study. ${\ddagger}$ denotes the full AS module (SC + SS sub-modules, using $\hat{v}_i$); ${\dagger}$ denotes using only the SC sub-module (using $\overline{v}_i$). The last two rows compare WMAPE vs.\ MSE as training loss.}
    %\vspace{0.2cm}
    \label{tab:ablation}
    % \begin{tabular}{c|c|p{20pt}p{20pt}|cc|cc|cc|cc|cc}
    \begin{tabular}{c|c|cc|cc}
        \hline
        \multirow{2}{*}{\shortstack{Methods/\\Datasets}} & &  \multicolumn{2}{c|}{Autoformer$_{AS}^{\ddagger}$} &  \multicolumn{2}{c}{Autoformer$_{AS}^{\dagger}$} \\
         & & WMAPE & RSE & WMAPE & RSE     \\ 
        \midrule[1pt]
        \multirow{2}{*}{Fund$_1$} &5 &95.18 &61.14&\textbf{93.27} &\textbf{60.38} \\
         & 10 &94.95 &61.79&\textbf{93.78} &\textbf{61.32} \\
        \hline
        \multirow{2}{*}{Fund$_2$} &5 &\textbf{86.33} &\textbf{59.93}&88.71 &61.10 \\
         & 10 &90.25 &62.35&\textbf{88.96} &\textbf{61.46}  \\
        \hline

        \multirow{2}{*}{\shortstack{Methods/\\Datasets}} & &  \multicolumn{2}{c|}{Transformer$_{AS}^{\ddagger}$} &  \multicolumn{2}{c}{Transformer$_{AS}^{\dagger}$} \\
         & & WMAPE & RSE & WMAPE & RSE     \\ 
        \midrule[1pt]
        \multirow{2}{*}{Fund$_1$} &5 &91.75 &\textbf{62.27}&\textbf{91.09} &63.51 \\
         & 10 &\textbf{94.64} &64.08&95.05 &\textbf{62.42}  \\
        \hline
        \multirow{2}{*}{Fund$_2$} &5 &\textbf{80.12} &\textbf{59.81}&88.19 &66.75 \\
         & 10 &\textbf{85.50} &\textbf{62.05}&90.27 &65.90  \\
        \hline
        \multirow{2}{*}{\shortstack{Methods/\\Datasets}} & &  \multicolumn{2}{c|}{Performer$_{AS}^{\ddagger}$} &  \multicolumn{2}{c}{Performer$_{AS}^{\dagger}$} \\
         & & WMAPE & RSE & WMAPE & RSE     \\ 
        \midrule[1pt]
        \multirow{2}{*}{Fund$_1$} &5 &\textbf{89.12} &\textbf{59.55}&90.27 &61.29 \\
         & 10 &\textbf{93.15} &\textbf{60.82}&95.57 &63.24  \\
        \hline
        \multirow{2}{*}{Fund$_2$} &5 &\textbf{79.18} &\textbf{58.23}&85.07 &64.52 \\
         & 10 &\textbf{86.13} &\textbf{61.36}&91.41 &66.20  \\
        \hline
        \multirow{2}{*}{\shortstack{Methods/\\Datasets}} & &  \multicolumn{2}{c|}{Informer$_{AS}^{\ddagger}$} &  \multicolumn{2}{c}{Informer$_{AS}^{\dagger}$} \\
         & & WMAPE & RSE & WMAPE & RSE     \\ 
        \midrule[1pt]
        \multirow{2}{*}{Fund$_1$} &5 &\textbf{83.07} &\textbf{59.43}&84.86 &62.57 \\
         & 10 &\textbf{94.01} &\textbf{64.08}&95.98 &66.75  \\
        \hline
        \multirow{2}{*}{Fund$_2$} &5 &\textbf{75.56} &\textbf{57.19}&80.98 &64.55 \\
         & 10 &\textbf{83.73} &\textbf{61.26}&88.74 &67.31  \\
        \hline

        \multirow{2}{*}{\shortstack{Methods/\\Datasets}} & &  \multicolumn{2}{c|}{Performer-WMAPE} &  \multicolumn{2}{c}{Performer-MSE} \\
         & & PMAE & RSE & PMAE & RSE     \\ 
        \midrule[1pt]
        \multirow{1}{*}{Fund$_1$} 
         & 10 &\textbf{45.79} &\textbf{61.97}&49.79 &63.75  \\
        \hline
        \multirow{1}{*}{Fund$_2$} 
         & 10 &\textbf{38.00} &\textbf{61.11}&42.52 &64.56  \\
        \hline
    \end{tabular}
%\vspace{-0.4cm}
\end{table}
%\vspace{-0.3cm}
\subsection{Main Results}
%\vspace{-0.1cm}
\textbf{Effectiveness of our AS module.} Table~\ref{tab:main_res} shows a clear performance hierarchy across scale-handling strategies. Models equipped with our AS module consistently outperform vanilla scaling (VS), which in turn outperforms no preprocessing (nop). This is because AS learns adaptive scale factors that better preserve semantic discriminability across different scales while reducing inverse-scaling restoration errors, compared to vanilla scaling that uniformly maps all windows to around 1. Furthermore, traditional global standardization (std) and normalization (nor) perform even worse than no preprocessing, confirming that these methods are unsuitable for scale-heterogeneous TSF. Our AS module achieves the best results across all four backbone models on both datasets, demonstrating its generality and effectiveness.

\textbf{Ablation study.} Table~\ref{tab:ablation} validates the contribution of each sub-module. On some models (e.g., Autoformer), using only the SC sub-module ($\dagger$, applying $\overline{v}_i$) already outperforms vanilla scaling, indicating that the learned calibration is effective. However, on other models (e.g., Transformer and Informer on Fund$_2$), SC alone over-calibrates certain well-performing prior factors, leading to degraded performance that even underperforms vanilla scaling. Adding the SS sub-module ($\ddagger$) resolves this issue: by selectively retaining well-performing prior factors and only calibrating the others, the full AS module consistently outperforms vanilla scaling across all settings. This confirms that SS effectively combines the strengths of $v_i$ and $\overline{v}_i$ through adaptive factor selection, ensuring that the worst-case performance of AS remains close to vanilla scaling while achieving substantial improvements in the best case.

\textbf{Loss function.} The last two rows of Table~\ref{tab:ablation} compare WMAPE and MSE as training losses. WMAPE yields consistently better results because it is scale-invariant, whereas MSE is sensitive to residual scale differences after scaling, leading to unstable training dominated by high-scale samples. Moreover, we also find that jointly modeling scale-heterogeneous variables with similar patterns via a shared backbone outperforms the fully multivariate setting by a large margin. We omit detailed results due to space limits.
%\vspace{-0.4cm}

\section{Conclusion}
%\vspace{-0.3cm}
This paper proposes the self-Adaptive Scale-handling (AS) module for time series forecasting under scale heterogeneity. Experiments on real-world fund sales datasets validate that: (1) the Scale Calibrating sub-module effectively learns input-adaptive scale factors that outperform both global normalization and vanilla window-based scaling; (2) the Scaling Selection sub-module further improves robustness by preventing over-calibration on already well-suited prior mean scaling factors; and (3) the scale-invariant WMAPE loss is better suited than MSE for scale-heterogeneous settings. The AS module is architecture-agnostic and consistently improves representative Transformer-based backbones. Future work involves extending AS to broader industrial scenarios with scale heterogeneity.

\section{Limitations}
Our current experiments involve two variables per product (purchase and redemption volumes). When extending to more variables with greater scale diversity within a single product, the learned calibration factors may become less stable or effective, requiring further exploration of cross-variable scale coordination. Additionally, the per-sample calibration factor can exhibit instability during training, and the current batch-mean strategy only partially alleviates this issue. More robust adaptive factor estimation methods remain to be investigated.

\bibliographystyle{IEEEbib}
\bibliography{main}

\end{document}